\pdfoutput=1

\documentclass[11pt]{article}

\usepackage{todonotes}
\usepackage{xspace}
\usepackage{color}
\usepackage{tipa}

\usepackage{acl}

\usepackage{times}
\usepackage{latexsym}

\usepackage[T1]{fontenc}

\usepackage[utf8]{inputenc}

\usepackage{microtype}

\usepackage{inconsolata}

\usepackage{graphicx}

\title{Multilingual MFA: Forced Alignment on Low-Resource Related Languages}

\author{Alessio Tosolini \\
  Yale University \\
  Department of Linguistics \\
  \texttt{alessio.tosolini@yale.edu} \\\And
  Claire Bowern \\
  Yale University \\
  Department of Linguistics \\
  \texttt{claire.bowern@yale.edu} \\}

\begin{document}
\maketitle
\begin{abstract}
We compare the outcomes of multilingual and crosslingual training for related and unrelated Australian languages with similar phonological inventories. We use the Montreal Forced Aligner to train acoustic models from scratch and adapt a large English model, evaluating results against seen data, unseen data (seen language), and unseen data and language. Results indicate benefits of adapting the English baseline model for previously unseen languages.
\end{abstract}

\section{Introduction}

Forced Alignment (the matching of textual annotations with audio and/or video data, particularly at the level of phonological segments) is a very useful step in language analysis. Software such as ELAN \citep{wittenburgELANProfessionalFramework2006} allows straightforward (but mostly manual) transcription and alignment at the granularity of utterances. Alignment algorithms such as the Montreal Forced Aligner \citep{mcauliffeMontrealForcedAligner2017} take utterances and align them at the level of words and segments, allowing a much greater array of analytical possibilities.

Forced Alignment requires an acoustic model and information about the mapping between the transcription system and the phonemes in the language (g2p). Acoustic models require training data, and the paucity of available materials for low-resource languages leads to lower model performance. Low-resource language materials are disproportionately created in naturalistic environments (outside quiet, controlled lab settings) and so in addition to having smaller amounts of data, the data that is there may be disproportionately difficult to work with.

Various methods exist for increasing performance, including a) using a very high resource language (mostly English) and adapting phoneme mappings to the high resource language; b) adapting a high-resource language model; c) using a closely related high-resource language model; d) using pretrained spoken term detection to identify particular words \citep{sanLeveragingPreTrainedRepresentations2021};
or e) training a language-specific model despite small amounts of data and correcting manually. \citet{chodroffComparingLanguagespecificCrosslanguage2024} compared these techniques and found that for small amounts of data (under approximately 25 minutes for their Urum and Evenki datasets), large cross-language and language-specific acoustic models were effective, but where the amount of low-resource data is larger than about 25 minutes, a model trained on that data is as effective. Findings by \citealt{sanPredictingPositiveTransfer2024} show that crosslingual transfer from models, as one might expect, is more effective when the languages are phonologically similar.  

For forced aligning Australian Indigenous corpus data, however, the question is somewhat different. In this case, we have a large number of phonologically similar \citep{roundSegmentInventories2023} but small corpora, which vary by number of contributors, circumstances and dates of recording, and language phonotactics. Since the languages are phonologically (and perhaps phonetically; \citealt{koch-3-2014,tabainStressEffectsStop2016}) similar, pooling data should lead to more robust and accurate alignment models. Conversely, since the languages differ in phonotactics \citep{macklin-cordesPhylogeneticSignalPhonotactics2021} and comprise different speakers, the increase in heterogeneity may limit improvements in model performance. Moreover, since even pooling data does not make the model ``large'' by ``large corpus'' standards, it may still be preferable to use or adapt a large model. 

For small corpora, overfitting is seldom a problem; model performance on the data at hand is often the sole criterion. In this case, however, we care about performance increases on both held-out data and held-out languages, as we will continue to develop the corpus and hope the release models for others working with Australian language data.

\begin{table*}[t!]
    \centering
    \begin{tabular}{lllp{4.5cm}c}
    \hline
      Language & Language Family & Reference   & Collector & Minutes \\ \hline
      Bardi & Nyulnyulan & A: Bowern\underline{ }C05 & Claire Bowern  & 108 \\
      Gija & Jarrakan & E: 0098MDP0190 & Frances Kofod & 157 \\
      Kunbarlang & Gunwinyguan & E: 0384SG0324 & Isabel O'Keefe; Ruth Singer & 16\\
      Ngaanyatjarra & Pama-Nyungan & P: WDVA1 & Inge Kral & 53 \\
      Yan-nhangu & Pama-Nyungan & E: dk0046 & Claire Bowern & 290\\
      Yidiny & Pama-Nyungan & A: A2616 & R.M.W.\ Dixon & 50\\
  \hline
    \end{tabular}
    \caption{Corpus information. A: AIATSIS; E: Elar; P: Paradisec}
    \label{tab:corpus}
\end{table*}

In this paper, we describe results of model training and evaluation for 5 MFA acoustic models. While previous work \citep{dicanioUsingAutomaticAlignment2013,johnsonForcedAlignmentUnderstudied2018,babinskiRobinHoodApproach2019a} has compared different alignment methods, here \citep[as in][]{chodroffComparingLanguagespecificCrosslanguage2024} we focus on comparing different acoustic language models within MFA.

We find general agreement between all but the model trained on the smallest amount of data. Adapting the English model for a crosslingual Australian dataset improves performance on held-out languages more than for held-out data from languages already in the dataset. Measurements of vowel space are equivalent for all except the smallest model when applied to seen languages, but there is more variation when applying models to a new language. This suggests that  similarity among Australian language phonetics should be further investigated.




\section{Methods and Data}

\subsection{Datasets}

Datasets for this paper were downloaded from non-restricted collections in the \texttt{ELAR}\footnote{\texttt{www.elararchive.org}} and \texttt{Paradisec}\footnote{\texttt{www.paradisec.org.au}} digital language archives, along with materials previously received from AIATSIS.\footnote{\texttt{mura.aiatsis.gov.au}} These materials are a subset of the collections which were used in \citet{babinskiArchivalPhoneticsProsodic2022}. Corpus references are in Table~\ref{tab:corpus}. The total amount of training data for the current study is roughly 10 hours, with individual languages ranging from 15 minutes to nearly 5 hours of audio. 

Some of the materials used here were initially used to compare forced alignment algorithms in \citet{babinskiRobinHoodApproach2019a}, and the full cleaned dataset was used for \citet{babinskiArchivalPhoneticsProsodic2022}. The data pipeline involved word-level segmentation with the \texttt{p2fa} forced alignment suite (based on HTK) and subsequent manual correction in Praat \citep{praat}. Manual correction included realigning substantially misaligned segments (for example, segment boundaries placed in the wrong word) and moving boundaries placed where no human annotator would place them. In this paper, those manually reviewed files are the comparator against which we evaluate the accuracy of the force aligned files.
However, we acknowledge that it is misleading to claim that there exists a single correct boundary between two phones due to smooth transition between phones that results from overlap (e.g. \citealt{liberman1967perception}) and that even expert annotators vary in regard to where they place phone boundaries. Moreover, some segments (such as word-initial glottal stops) might not have any detectable onset boundary. We compare models to human annotated data but acknowledge that such datasets are themselves subject to further scrutiny.   

The languages that form the basis of this comparison do not have identical phoneme inventories. They differ as to whether they have phonemic vowel length (or not) and whether they have two series of stops or one. For languages with two stop series, the contrast is between voicing, length, or perhaps tense/laxness (or some combination of these features).

\subsection{Preparing Input}


Since the audio data collected for this experiment comes from a variety of sources, we preprocessed the data to standardize it and to ensure that (a) the data is processed as expected by the various MFA models we created and trained and (b) all datasets created have the same formatting. This processing included the removal of partially transcribed words, cleaning the transcription tier of analytical comments, and some transcript adaptation (such as the removal of hyphens). TextGrids processed for model evaluation underwent additional processing to match it with the expected output of the MFA model, such as removal of words shorter than 0.1s in duration. We did not alter transcripts.\footnote{These transcripts do not  mark pauses or hesitations in the original. We did not review transcripts for accuracy beyond what was completed for earlier publications.} 
Two datasets were created for model training. One of them is the Yidiny-Train corpus, comprised of 38 minutes of audio data. The other is the Big5 dataset, comprised of the entirety of the Bardi, Gija, Ngaanyatjarra, and YanNhangu corpus, and the Yidiny-Train corpus. The Big5 has a total duration of 646 minutes.
Three datasets were created for model evaluation. The first is the same as the Yidiny-Train corpus, and the second is the comprised of the remaining 12 minutes of Yidiny data the models didn't train on. The last test corpus is the Kunbarlang corpus, which no model has trained on.  

It must be noted that although similar, the phonemic inventories of these languages are different enough to play a significant role in the alignments generated. Notably, Kunbarlang's phones are not a proper subset of Yidiny's, with the tense stops /p t \textrtailt \hspace{0.04cm} c k/, mid-vowels /e o/, and the retroflex nasal /\textrtailn / all present in Kunbarlang but absent in Yidiny. All of Kunbarlang's phones are present in at least one of the models in the Big5 training set, with the exception of the mid front vowel /e/ which is not present in any of the Big5 languages. The impact of these inventory asymmetries is discussed in the section on results.

\subsection{Acoustic Models}

Five acoustic models were used for this experiment. The first two of these acoustic models were trained from scratch on the Yidiny-Train and Big5 corpora. The remaining three models use the English MFA 3.1.0 acoustic model \cite{mfa-english-mfa-acoustic-2024}, which is trained on over 3,600 hours of global English. In order to use the English models for non-English data, we used the methodology described in \citet{interlingual-mfa}. The English base model was used in its off-the-shelf form as one of the models we evaluated. The other two English-based models were created by adapting the English model to the Yidiny-Train and Big5 corpora. 

The dictionary was created from the corpora by creating a single wordlist of all language data to be included and replacing graphemes in the language orthographies with equivalents from the International Phonetic Alphabet. Since all languages used phonemic transcription systems this was straightforward.

\subsection{Eval}

We evaluated models against two criteria in three different testing settings. The first criterion is precision, defined as the distance in milliseconds between the human annotated onset boundary and the MFA aligned interval's onset boundary, where a positive value indicates the aligned onset boundary is placed after the human annotated onset boundary. To check for both accuracy and precision, we look at the mean and standard deviation for these values, which we call ``diffs". 

The second criterion is analysis comparison. Since the aligned output of a model is used for phonetic analysis, we compare vowel charts created from these model against those created by the human annotated files. To test for accuracy and precision, we plot ellipses centered around the mean formant values for the data using the \texttt{matplotlib} package in Python \citep{Hunter:2007}. Formants were extracted using  \texttt{Parselmouth}  \citep{parselmouth} and measurements averaged across each vowel.

The three testing settings correspond to the three human annotated datasets described in the previous subsection. They are: Yidiny-seen (comprised of Yidiny data the model has trained on), Yidiny-unseen (comprised of Yidiny data the model has not trained on), and Kunbarlang (comprised of Kunbarlang data). Note that all models except English-base have trained on some amount of Yidiny data, and that no models have trained on any Kunbarlang data. 

\section{Results} 
\subsection{Precision}
Figures in this section present the rules of mean differences in onset alignment of segments in milliseconds, where a positive value means that the forced-aligned onset boundary has a greater timestamp than the human annotated onset boundary (i.e. it is further on in the file).

\begin{figure*}
    \centering\vspace*{-4cm}
    \includegraphics[width=\linewidth]{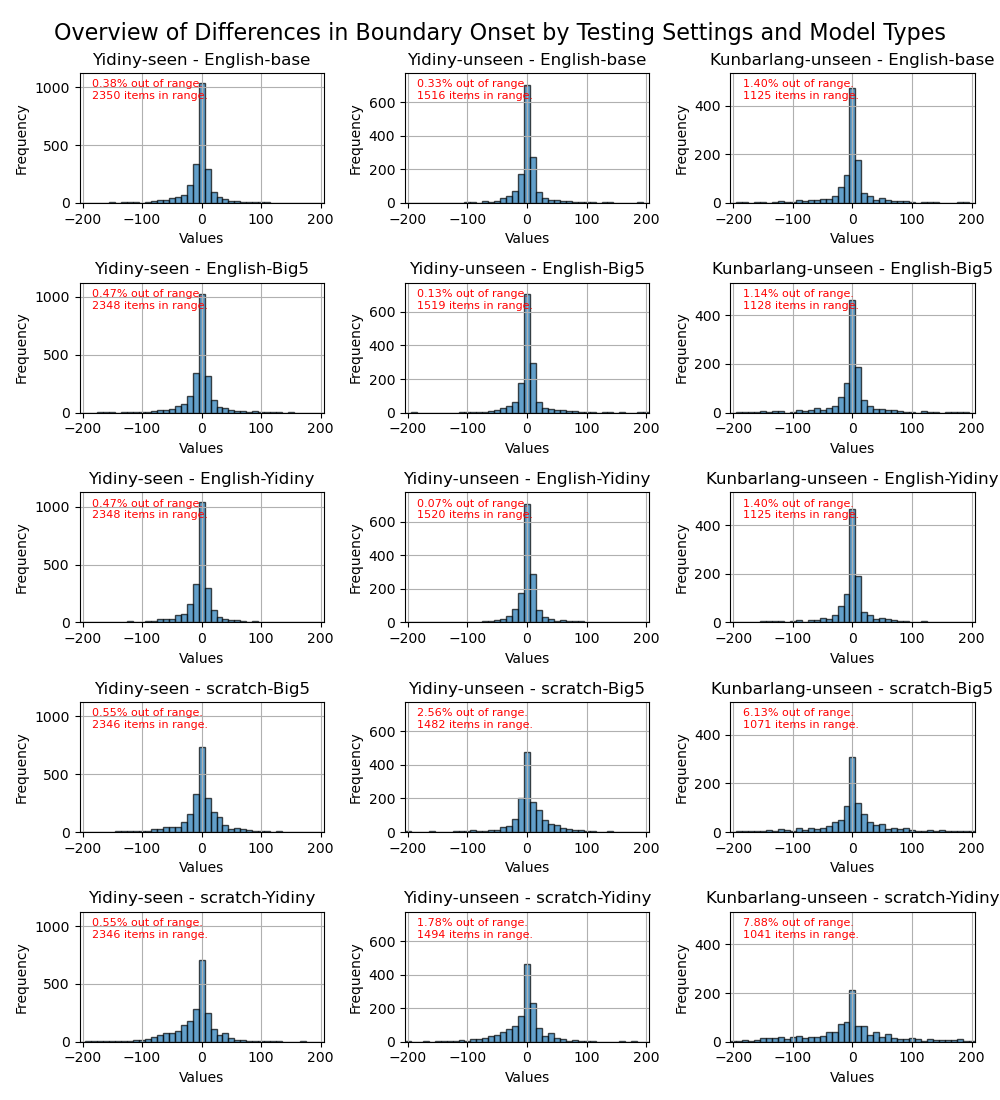}
    \caption{Onset boundary differences for all models across all testing settings.}
    \label{fig:ComparisonHistogram}
\end{figure*}

Figure~\ref{fig:ComparisonHistogram} shows 15 histograms, where each row is a different model and each column is a different testing setting. The histograms only plot values in the range of [-205, 205] milliseconds, with the percentage in the top left equal to the percent of total tokens that were excluded from the histograms due to being out of this range. The number below represents the test tokens per testing setting that were in the range.

From this, we see that all diffs are approximately normally distributed with a mean near 0. Models trained from scratch have more spread, which is especially notable with the lower number of boundaries that differ from the human annotated boundaries by [-5, 5] ms. Although the histograms seem roughly symmetrical, there is tendency for English-based models tested on unseen data to place boundaries slightly ahead of the human annotated ones, as is seen by the higher number of tokens falling in the [5, 15] bin than the [-15, -5] bin for those models.

\begin{figure}
    \centering
    \includegraphics[width=\linewidth]{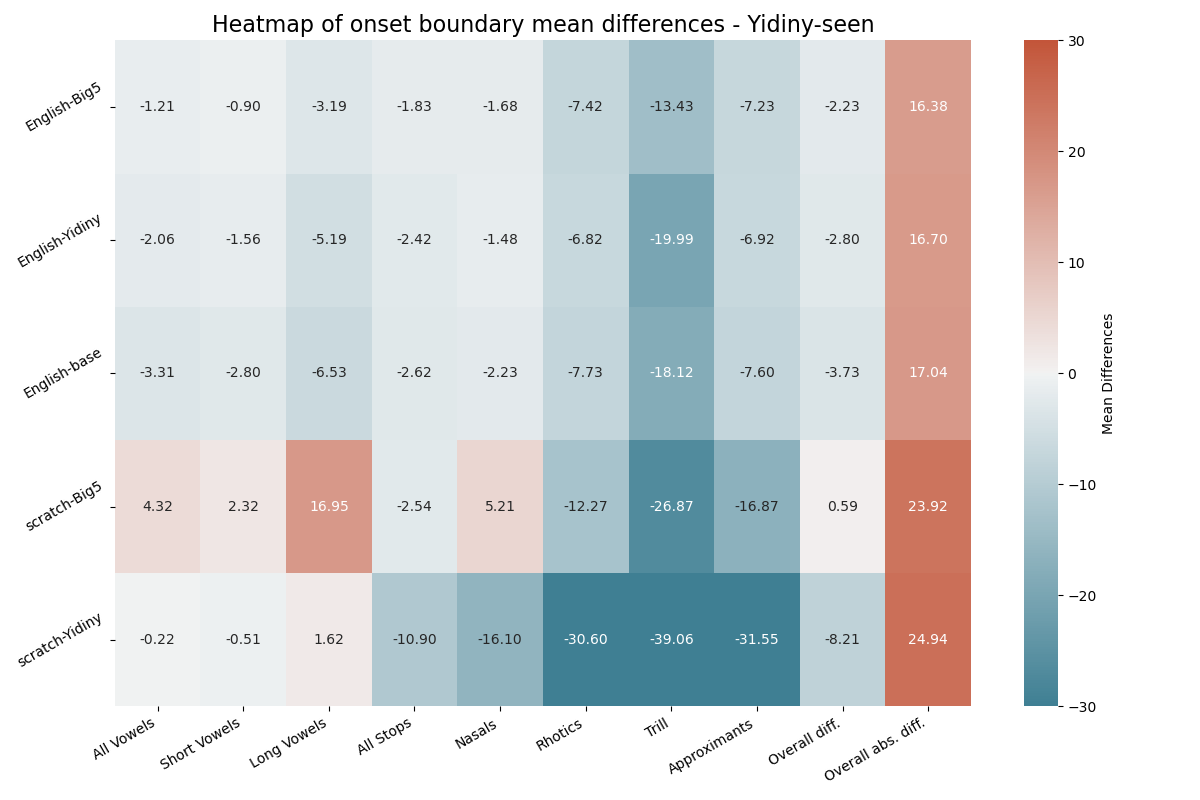}
    \caption{Yidiny  seen data; mean precision}
    \label{fig:YidinySeenMeans}
\end{figure}

Figure~\ref{fig:YidinySeenMeans} gives the results for seen language data. In this condition, the best performing model is the English model adapted to other Australian languages; however, adaptation only gives marginal improvements compared to the unadapted English model. Unsurprisingly, the best gains arise from segments which are not well represented in the English data (trilled rhotics, IPA /r/), while the gains over using a model trained only on Australian languages are those segments which are rare (long vowels) or difficult to identify boundaries for (approximants).

\begin{figure}
    \centering
    \includegraphics[width=1\linewidth]{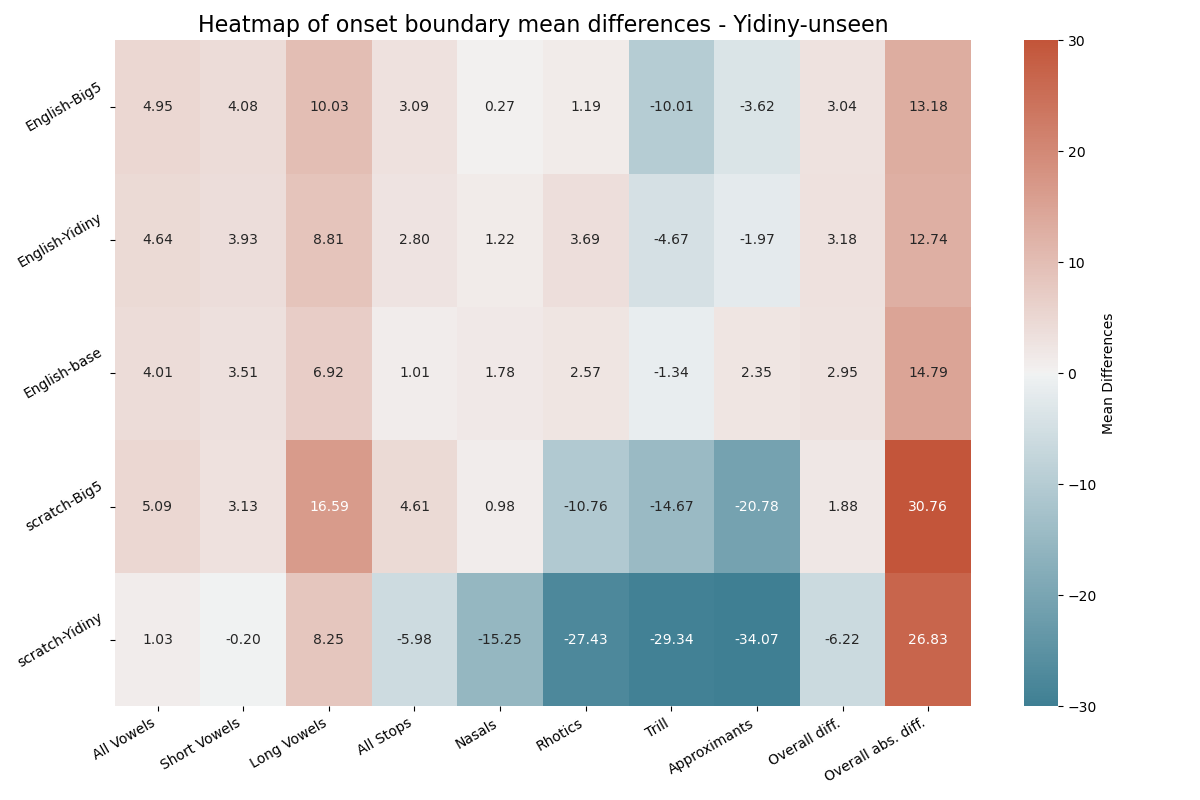}
    \caption{Yidiny unseen data; mean precision}
    \label{fig:YidinyUnseenMeans}
\end{figure}

For held-out Yidiny data, results are similar (see Figure~\ref{fig:YidinyUnseenMeans}). Here are there larger gains from adapting, but the adapted model does worse than the unadapted one on trills and long vowels. This might imply that there are characteristics of individual audio files that are affecting the results (we made no attempt to control for constant background noise, for example). Interestingly, the mean absolute diffs across only the models trained from scratch is greater for testing on unseen data than seen data.

\begin{figure}
    \centering
    \includegraphics[width=\linewidth]{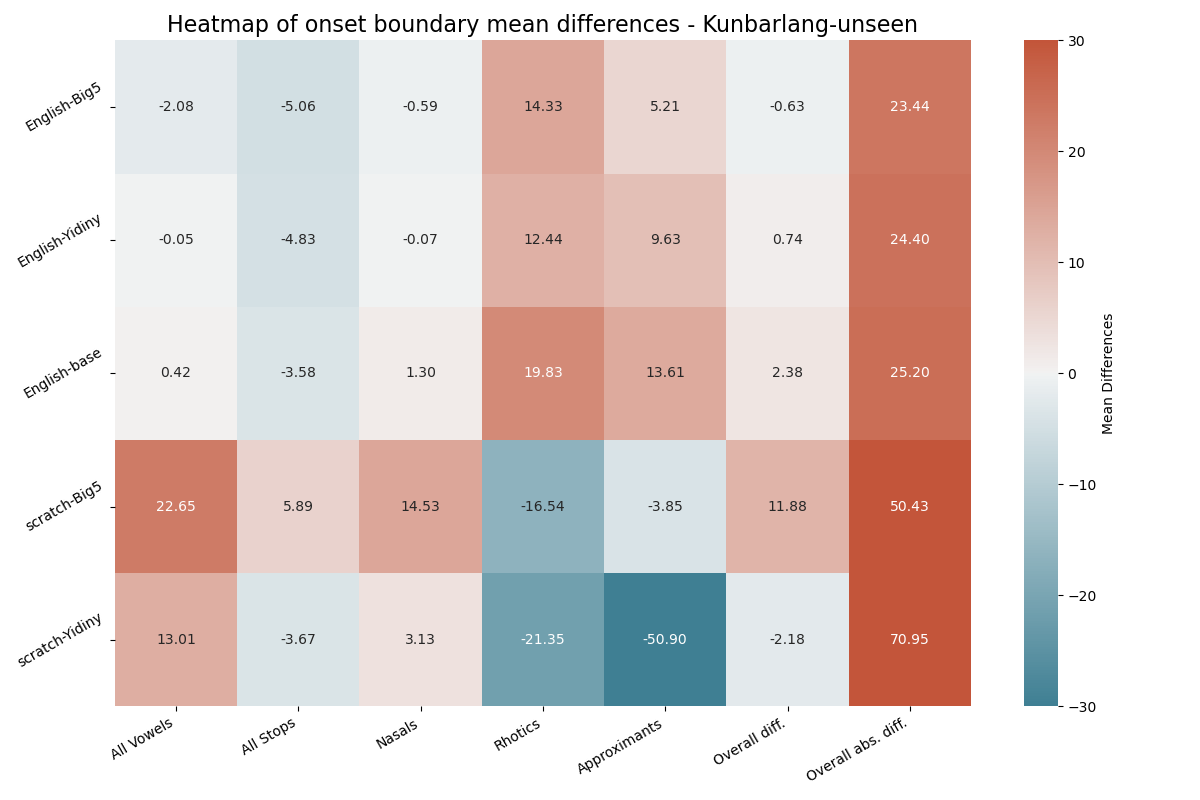}
    \caption{Kunbarlang (unseen), mean precision}
    \label{fig:KunbarlangUnseenMeans}
\end{figure}

Fig.~\ref{fig:KunbarlangUnseenMeans} shows results for Kunbarlang, a language that no model trained on. Overall, all models perform worse on Kunbarlang than Yidiny data in either setting. For Kunbarlang rhotics and approximants, English-based models consistently predict the boundary is ahead of the human annotated boundaries while from-scratch models consistently predict the opposite. The Kunbarlang setting is also the setting where we see the greatest difference in the accuracy of the models trained from scratch on the Big5 dataset and the Yidiny dataset. This is not surprising, as there are many phones in Kunbarlang which are not present in Yidiny but are present in one of the Big5 languages. 

Models trained from scratch on multilingual Australian data do very poorly on held out data, implying, perhaps, that there is not as much similarity between Australian languages as has been previously asserted, or that at least models are not able to take advantage of the similarities that do exist between languages.

\begin{figure}
    \centering
    \includegraphics[width=\linewidth]{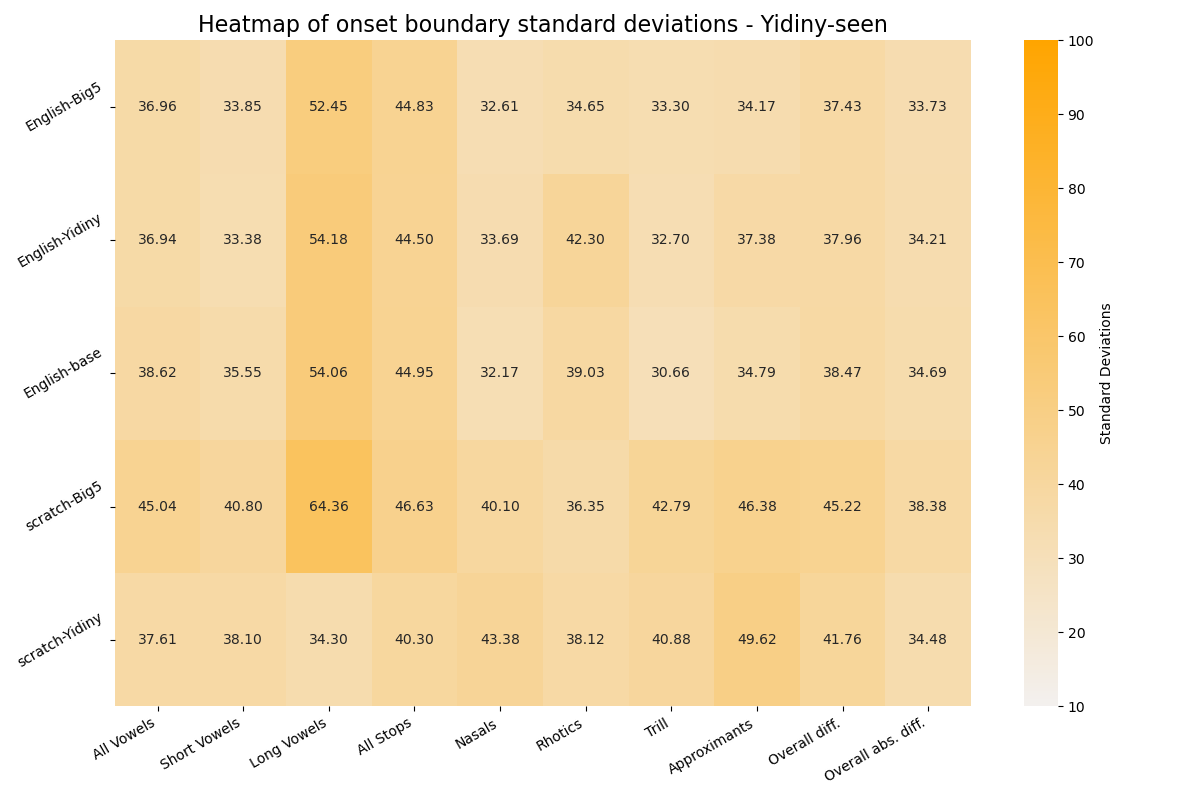}
    \caption{Yidiny (seen language) seen data, standard deviations of precision}
    \label{fig:YidinySeenStDs}
\end{figure}

Since a model with high accuracy and low precision would give an illusion of excellent performance, heatmaps for the standard deviation of onset boundary per natural class is provided below. Figure~\ref{fig:YidinySeenStDs} shows the standard deviation of the diffs for models tested on seen Yidiny data.

In this condition, the most consistent model is again the English model trained on the data from 5 languages. Perhaps surprising however, is the comparable precision of all 5 models. The English based models also differ from the models trained from scratch in the natural class of phones that they align most precisely, with the English models' trill and approximant onset boundaries differences with the human annotated data having a lower standard deviation than the models trained from scratch. 

\begin{figure}
    \centering
    \includegraphics[width=\linewidth]{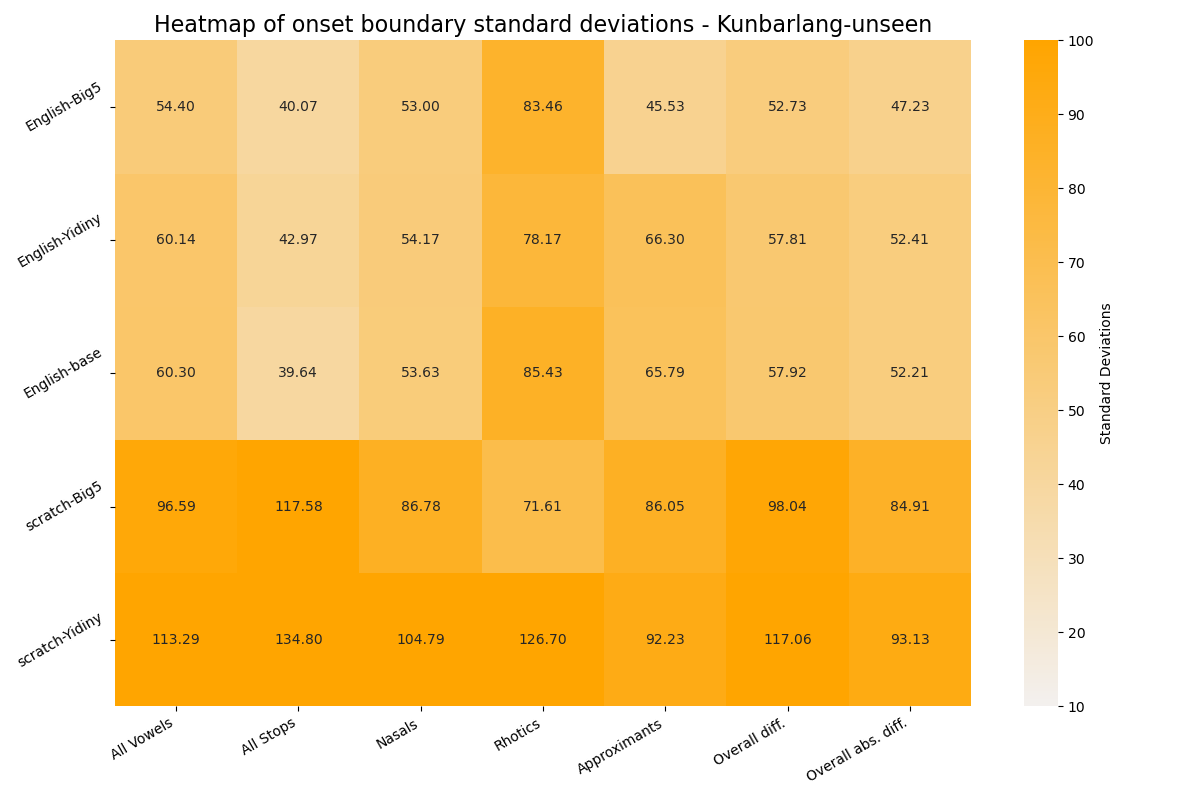}
    \caption{Kunbarlang (unseen language) unseen data, standard deviations of precision}
    \label{fig:KunbarlangUnseenStDs}
\end{figure}

As seen in Figure~\ref{fig:KunbarlangUnseenStDs} testing on unseen language Kunbarlang, we find that all English models give more precise onset boundaries than their from-scratch counterparts fairly independently of the natural class of the phone. Adapting an English model also gives more precise measurements than the base English model, although for both English-adapted and from-scratch models, attempting to augment the training data with data from related languages lowers precision.

\begin{figure}
    \centering
    \includegraphics[width=\linewidth]{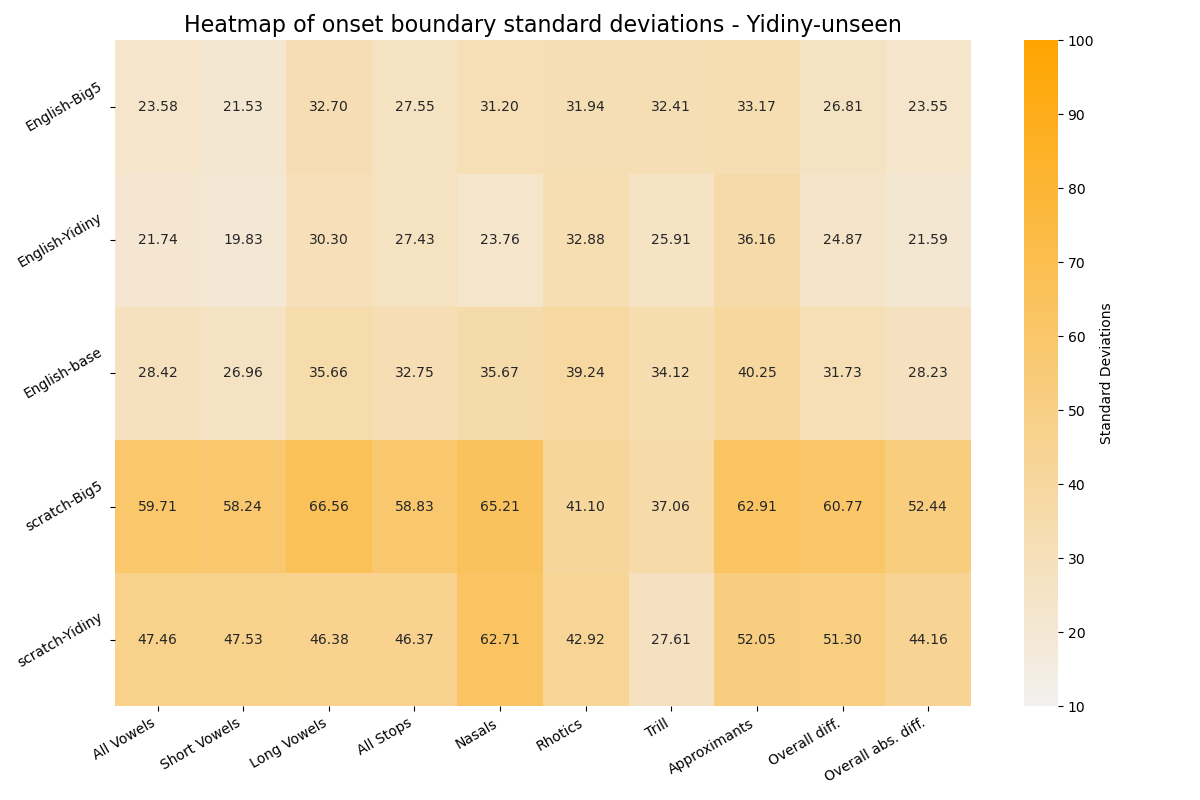}
    \caption{Yidiny (seen language) unseen data, standard deviations of precision}
    \label{fig:YidinyUnseenStDs}
\end{figure}

The biggest differences between the precision of the from-scratch and English-adapted models occurs for a language that was not in the training data (see Figure~\ref{fig:KunbarlangUnseenStDs}). In the unseen language setting, training on the Big5 dataset results in a notable improvement in precision in precision compared to training only on Yidiny data.

\subsection{Analysis Comparison}
As one might expect, given the overall similarity in precision of boundary identification discussed above, vowel dispersion plots show minimal differences between models. The exception is the model trained from scratch on a single Australian language which consistently has noticeably different vowel ellipses from the other from-scratch model and the English-based models. 

For all plots, a character representing the standard IPA transcription for the vowel quality is placed at the mean F1, F2 of the vowel, and ellipses are drawn with the horizontal and vertical axes of the ellipse representing two times the standard deviation of F2 and two times the standard deviation of F1 respectively. The color of the character and the ellipse corresponds to the model used to generate alignment. Black solid lines represent the plots made with formant values extracted from the human annotated files.

\begin{figure}
    \centering
    \includegraphics[width=1\linewidth]{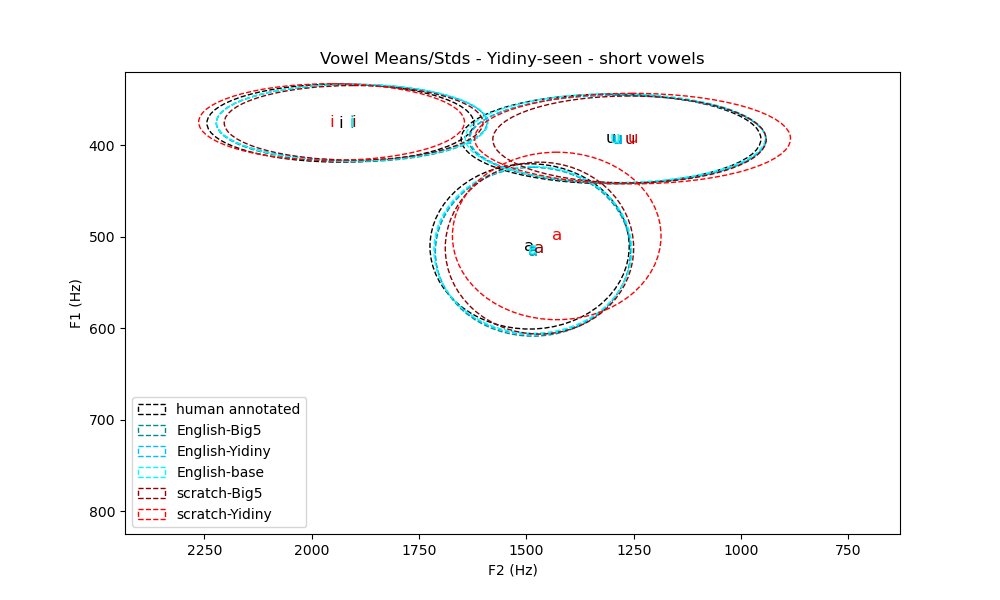}
    \caption{Comparison of vowel space measurements (F2:F1), short vowels, Yidiny seen data}
    \label{fig:YidinySeenShortVowels}
\end{figure}

Figure~\ref{fig:YidinySeenShortVowels} shows vowel plots for the three short vowels in Yidiny. There is much similarity in the ellipses and mean values for each value, with the exception of the from-scratch model trained only on Yidiny data. All English-based models and the from-scratch model trained on more data produce analyses with similar ellipses as the human annotated files. The same trend of highly accurate means and ellipses can be seen with the short vowels in the Yidiny-unseen testing setting (see Figure~\ref{fig:YidinySeenShortVowels}). Again, the only model which seems to produce notably incorrect results is the Yidiny-only model.

\begin{figure}
    \centering
    \includegraphics[width=1\linewidth]{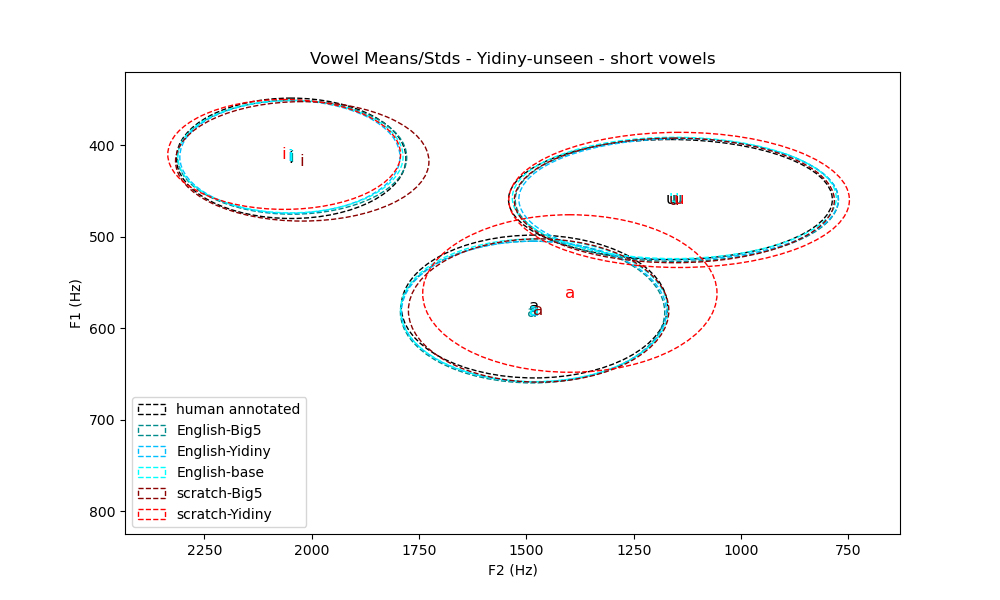}
    \caption{Comparison of vowel space measurements (F2:F1), short vowels, Yidiny unseen data}
    \label{fig:YidinyUnseenShortVowels}
\end{figure}

\begin{figure}
    \centering
    \includegraphics[width=1\linewidth]{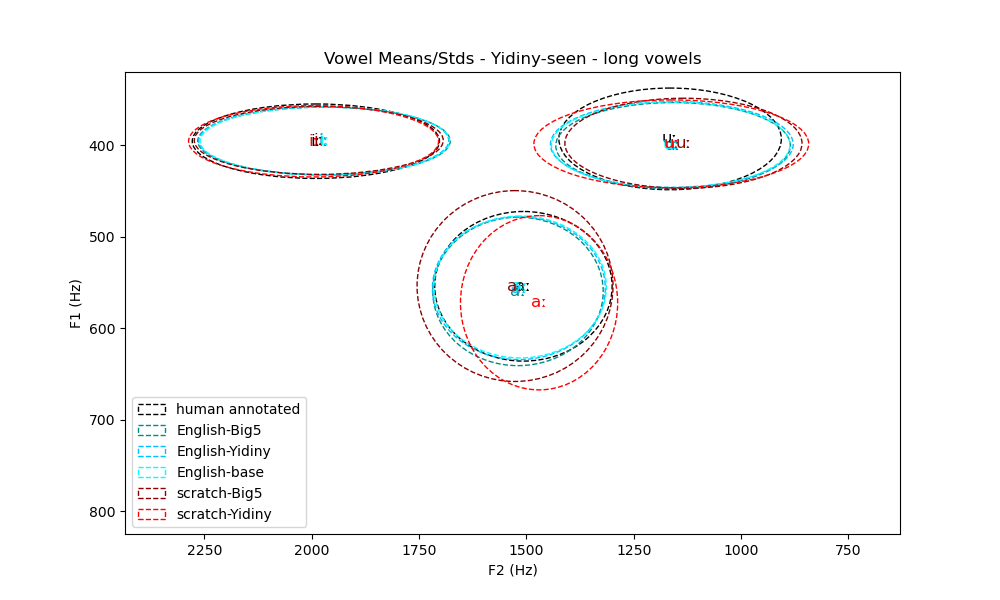}
    \caption{Comparison of vowel space measurements (F2:F1), long vowels, Yidiny seen data}
    \label{fig:YidinySeenLongVowels}
\end{figure}

There is more variation in the analyses of long vowels than short vowels, as is seen in Fig.~\ref{fig:YidinySeenLongVowels}. The tendency for the model trained on five Australian languages from scratch to perform similarly to the English-based models is no longer observed, with the ellipses of long vowels being not only larger than the English-based models but also larger than the model trained from scratch on less data. Again, the predictions of English-based models are nearly identical to those derived from human annotated data.

\begin{figure}
    \centering
    \includegraphics[width=1\linewidth]{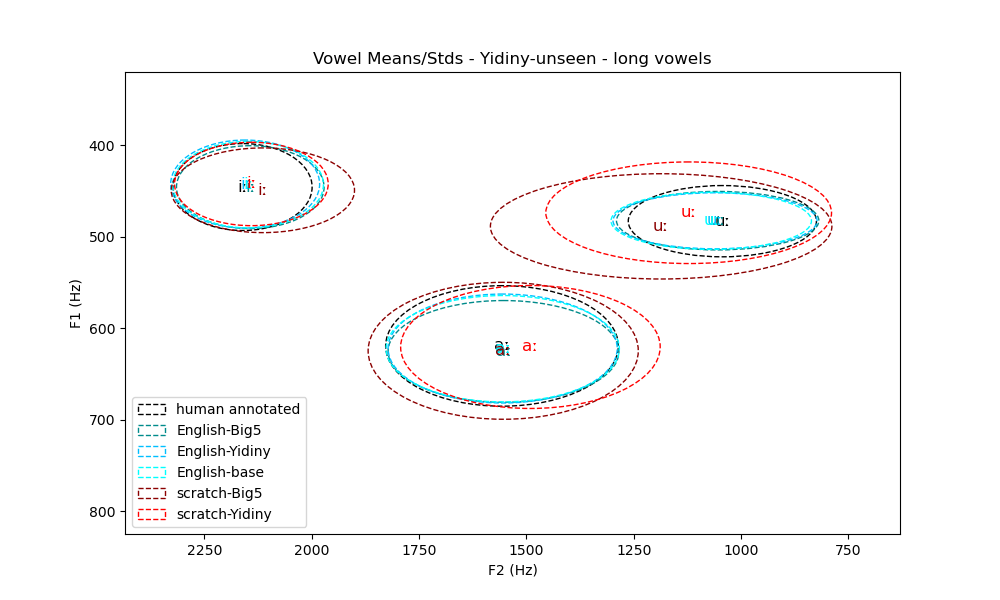}
    \caption{Comparison of vowel space measurements (F2:F1), long vowels, Yidiny unseen data}
    \label{fig:YidinyUnseenLongVowels}
\end{figure}

The relationship between Yidiny-unseen short and long vowels mirrors that between Yidiny-seen short and long vowels (see Figure~\ref{fig:YidinyUnseenLongVowels}). English-based models give similar vowel analyses to the human annotated data, but the models trained from scratch are noticeably inaccurate, with ellipses that are notably larger than the ellipses generated from human annotated data. 

Kunbarlang has a five vowel system that does not contrast for vowel length. No model provides an analysis almost identical to the human annotated standard, but all English-based models demonstrate high accuracy with means approximating the human annotated boundaries well. The model trained from scratch on Yidiny is notably inaccurate for the mid and low vowels. The model trained from scratch on the Big5 dataset provides similarly inaccurate results for /e/, which is also not present in any Big5 language, but shows better results for /o/ which is present in Bardi. The models trained from scratch are imprecise in this setting, with ellipses that do not approximate the human annotated ellipse. The vowel ellipses of English-based models approximate the human annotated ellipses more closely than the from-scratch models.

\begin{figure}
    \centering
    \includegraphics[width=1\linewidth]{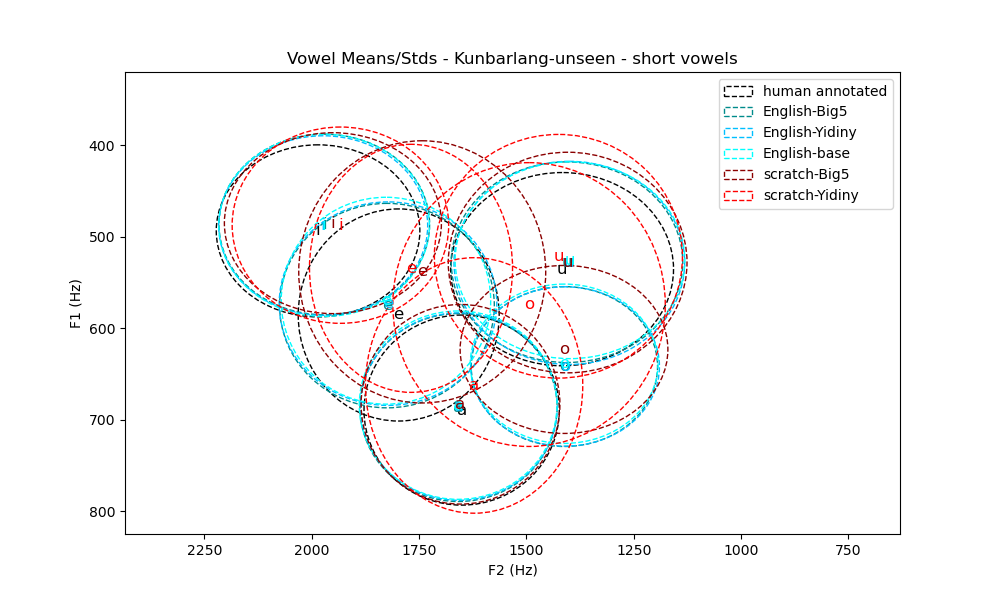}
    \caption{Comparison of vowel space measurements (F2:F1), short vowels, Kunbarlang (unseen language) vowel ellipses}
    \label{fig:KunbarlangVowelMeans}
\end{figure}

\subsection{Further Comments on Errors}

\begin{figure}
    \centering
    \includegraphics[width=0.5\linewidth]{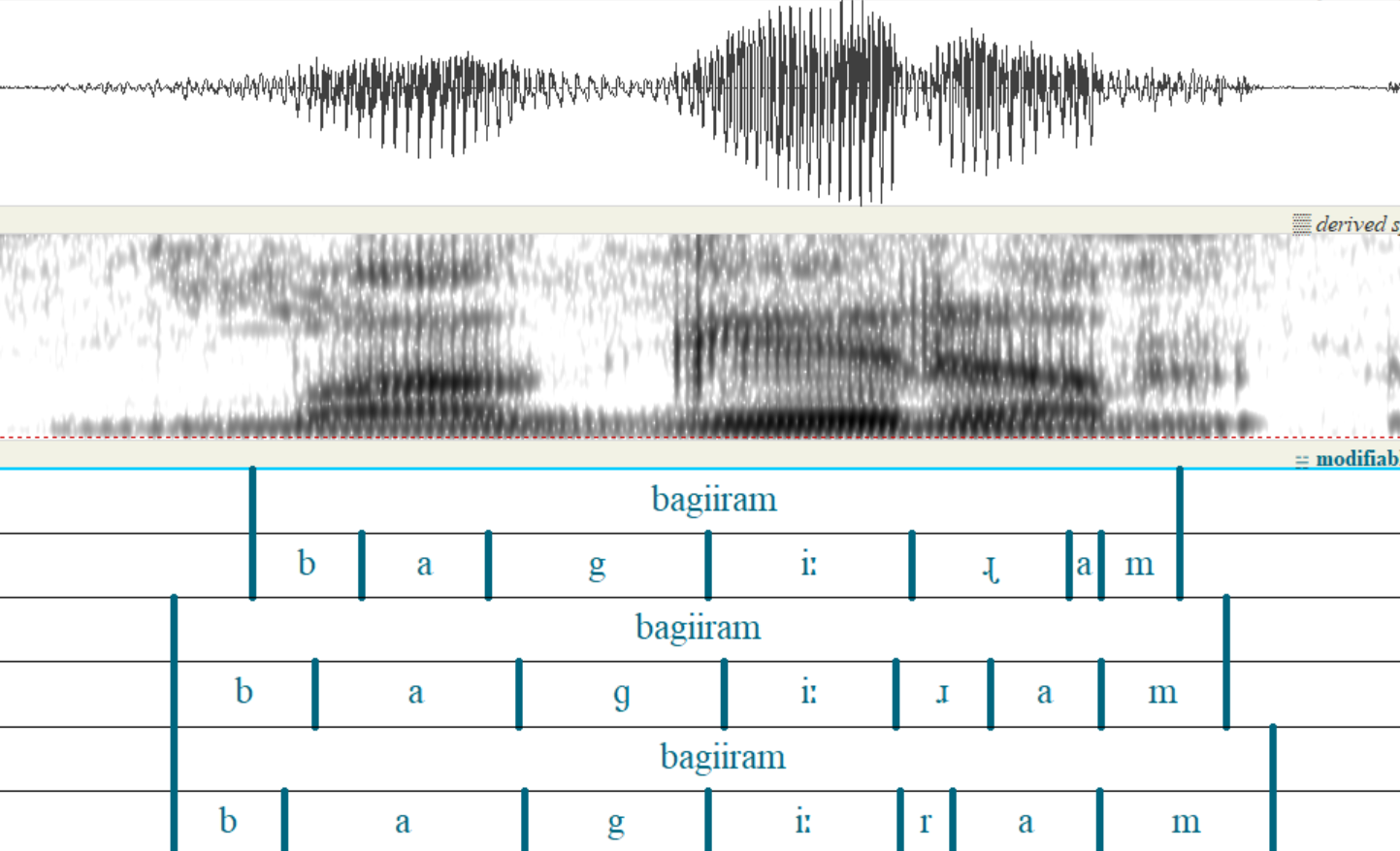}
    \caption{Typical set of textgrids of Yidiny seen data, with the scratch(big5) (top), English-adapted(big5) (middle), and human corrected (bottom) textgrids, illustrating errors in alignment.}
    \label{yidinydata}
\end{figure}

In order to further investigate common types of errors, we manually compared and spot-checked alignments in Praat. Figure~\ref{yidinydata} shows an example textgrid of Yidiny seen data, with the scratch(big5) (top), English-adapted(big5) (middle), and human corrected (bottom) textgrids placed on top of one another for comparison. To investigate the major sources of misalignments, we tagged the first 100 items in the file that varied from human-annotated data by more than 100 ms. Errors in the test (Yidiny-unseen) data are similar in kind and relative frequency, but more copious. While human annotators may differ in the placement of boundaries in continuous data, the mismatches studied here are considered errors because for these cases, the alignment models place boundaries in areas where no human annotator would do so.

One third of the tagged errors arose from difficulties in identifying stop boundaries: onsets to stops in initial position, or onset or offset of closures in medial position. Yidiny stops (as has been reported for other Australian languages) can have debuccalized realizations, with extensive lenition and no clear closure or release burst (see, for example, \citealt{enneverReplicableAcousticMeasure2017a}). Almost another third arise from nasal boundaries in medial or final position. Most of the rest of the errors come from intervocalic rhotic, lateral, or glide identification vis \`a vis adjacent vowels. Such errors (except for those with initial and final segments) affect measurements of surrounding segments; 2/3 of the errors involved word-internal segments.

\section{Discussion and Conclusion}

Overall, we find that the most accurate models across all testing settings were the models with the global English model as a base. For seen data, English-based models slightly outperformed models in mean diff compared to models trained from scratch. However, when aligning unseen data from a seen language, English-based models produced mean diffs equal to about half of their from-scratch counterparts. This is consistent with the robustness of English models, a result of training on extremely large amounts of data. For the unseen language setting, English-based models have about half and a third of the mean absolute diff than the multilingual and monolingual  models trained from scratch respectively. These findings suggest that English-based models are consistently more accurate than models trained from scratch in settings where there is little to no data for the target language.

In terms of adapting, we find that adapting the English-base model on the Big5 corpus provided marginal improvements for the Yidiny-seen and Kunbarlang settings compared to adapting on only the Yidiny-train corpus, but not for the Yidiny-unseen setting, suggesting that adapting on more data from related languages might ``dilute'' the effects of training on the language being tested on. All models struggled with rhotics, trills, and approximants, which is probably a result of the lack of good correspondences for the rhotics and trills present in Australian languages and a lack of a clear transition from the onset and offset for sounds belonging to these natural classes. However, across all settings the improvements of adapting an English-based model are marginal.

Of the three testing settings, we find that training a model from scratch on a multilingual dataset provided a notable 29\% improvement when testing on a language that the models have never seen before. This fits with the intuition that a model trained on more languages has more flexible representations for what each phone may look like, and is thus better able to leverage that knowledge in a new setting. This effect is much more noticeable in situations where a phone in the testing language is not present in the monolingual dataset but is present in at least one language from the multilingual dataset. This is exemplified with the vowel plots on Kunbarlang data, where both models trained from scratch struggle with plotting /e/ to its absence in the training data but the Big5-trained model gives a much better analysis of /o/ due to its presence in Bardi. It should be noted that /o/ still appears infrequently in Bardi, with it being the least frequent vowel quality and the only one to lack a long counterpart.

The improvement from training on more multilingual data is minimal for the Yidiny-seen setting, and actually negative for the Yidiny-unseen setting, suggesting that more data from related languages won't necessarily increase performance when testing on seen data and may actually hinder performance when testing on unseen data from a seen language. This can be explained using the same logic of ``diluting'' the data described in the previous paragraph.

The above results are mirrored when looking at the vowel analyses produced by the alignments output by the various models. For all testing settings, the plots generated from the alignments from the English-based models closely resembles the ones generated by the human generated alignments. Multilingual models trained from scratch performed comparably to the English-based models for short vowels, but produced visibly more imprecise measurements for long vowels, possibly due to long vowels having less tokens for these models to train on.

Ultimately, models trained from scratch on low-resource languages  suffer from the small amount of data and the resulting lack of variety in training examples. Future research should explore whether there exist data augmentation methods that may alleviate data scarcity by providing a slightly acoustically modified version of the input audio, artificially increasing the amount of data a model sees during training. Additionally, overfitting is not an issue in most low-resource settings due to model performance on seen data being the most relevant metric for downstream tasks. Future research may thus explore the effects of hyperparameter tuning a model to encourage overfitting, sacrificing model generalizability for a performance boost on seen data.

The findings presented in this paper are useful in the context of language documentation and revitalization, because they highlight the effectiveness of using a pretrained global English model on field data. The availability of the global English pretrained models and ease of adapting them to other languages means that high quality forced alignment is accessible to any fieldworker. The similarity of the vowel plots for the Big5 models trained from scratch and the English-based models also show promise that medium-sized multilingual training datasets can provide a boost in low-resource setting.

\section*{Acknowledgements}
Thanks to the participants at ComputEl8 (colocated with ICLDC in Hawai`i) and to Yale's Phonology and Fieldwork groups for useful feedback.



\bibliography{references,custom}




\end{document}